\pdfoutput=1
\documentclass[runningheads]{llncs}

\usepackage{booktabs}
\usepackage{graphicx}
\usepackage{makecell}
\usepackage{xcolor}
\usepackage[autolanguage]{numprint}
\usepackage{amsmath}
\usepackage{algorithm}
\usepackage{algpseudocode}
\usepackage{hyperref}
\usepackage{orcidlink}

\newif\ifshowpurple
\showpurplefalse     %

\newcommand\npr[1]{}

\title{Letter Lemmatization: One-to-one and Banded RNNs for Reversing Character-Set Simplification and Abbreviation in Medieval Text}
\titlerunning{Letter Lemmatization: One-to-one and Banded RNNs}
\authorrunning{A. Nicolaou et al.}

\author{%
Anguelos Nicolaou\inst{1}\,\orcidlink{0000-0003-3818-8718} \and
Maria Pia Tiseo\inst{2,3} \and
Tam\'{a}s Kov\'{a}cs\inst{1}\,\orcidlink{0000-0002-3913-2946} \and
Nicolas Renet\inst{1} \and
Georg Vogeler\inst{1}\,\orcidlink{0000-0002-1726-1712}%
}
\institute{%
University of Graz, Graz, Austria\\
\email{\{firstname.lastname\}@uni-graz.at}
\and
University of Basel, Basel, Switzerland
\and
Universit\`a degli Studi di Napoli Federico II, Naples, Italy%
}

\begin{document}

\maketitle

\begin{abstract}
Medieval document transcribers have very different practices;
on top of that, heterogeneous digitization policies have resulted in corpora where the character-set must be viewed as fluid.
In this paper we address the problem of changing between character-sets in a flexible manner.
We focus on one-to-one character mappings and train character-level one-to-one RNNs to undo them with self-supervision; recovering half the CER even with 20 text lines.
We analyse the use of these one-to-one networks for HTR post-correction and we see that they obtain significant improvements while totally ignoring ins-dels.
We then use the exact same networks with character-level alignment groundtruth compiled from parallel corpora in a training and inference mode we call Banded RNNs.
We use such networks to successfully expand abbreviations in medieval charter transcriptions.
Finally we introduce an elaborate heuristic which takes the characters of two arbitrary character-sets and defines a metric encapsulating what we consider to be semantic similarity of characters.
We call the construction of such mappings letter lemmatization and present a rich Python library that efficiently performs all presented methods.
\end{abstract}

\section{Introduction}
In this paper, we address the question of which character set to use when digitizing medieval documents and how to effectively change between character sets.

While acknowledging normalization issues that predate the digital era~\cite{bollmann-2019-large}, as well as the perils of making transcriptions conform to an elusive, ideal state of a language in the late Middle Ages~\cite{kragl2015}, we do not intend to take a linguist's stance on these topics but rather to serve the pragmatic needs of the curators and remote readers of digital corpora.
Are there ways to align character sets that make it easier to train digitization tools and to evaluate their performance consistently, without erasing the variations over time and space that make these corpora relevant to the digital humanists in the first place?

Conflicting needs and tradeoffs make it impossible to have a clear answer to which approach is optimal, yet we try to provide quantifiable insights on the question.
Optical Character Recognition (OCR) has a long history, with open-source OCR tools driving research on the topic~\cite{smith2007overview,breuel2008ocropus,ocrd,kiessling2019kraken}.
With the introduction of Neural Networks (NN) for OCR, OCR performance increased~\cite{breuel2013high} to the extent that they could also perform usable Handwritten Text Recognition (HTR)~\cite{icdar2015_HTRcompetition,icdar2017_HTRcompetition}, a much harder task.
Unlike OCR, HTR cannot be trained on synthetic data effectively and is more sensitive to the handwriting style.
Many digitization projects nowadays employ HTR to digitize the vast majority of their corpora, but they usually need manual transcription of representative parts of their corpus to adapt HTR engines to their data and have a comprehensive quality control of the process.
Our approach assumes that HTR training and quality evaluation are a principal use-case for transcribing text.
At the same time, we recognize that linguistic analysis, distant reading~\cite{distantReading_moretti2000conjectures}, and automatic indexing of corpora are also important use cases and they are not all best served by the same character set when it comes to text encoding.
Character sets are also a key issue when it comes to harmonizing different corpora in order to combine them into larger ones.
A smaller, consistent character set further facilitates search and indexing, since a query and its targets no longer need to agree on the encoding of every variant glyph.

When dealing with medieval documents, as is our use case, the Medieval Unicode Font Initiative (MUFI)~\cite{haugen2009mufi} features the most extensive character set, but due to its ambitious size (1501 characters), it introduces many problems, such as 11 characters beyond Unicode's Basic Multilingual Plane (BMP) and 713 characters inside the Private Use Areas (PUA) of the BMP which are not considered part of the Unicode standard~\cite{unicode2021}.
Even the 777 characters of MUFI, which are outside the PUA and inside the BMP, may be too many for most use cases.
Most importantly, since HTR models essentially classify every symbol as one of a list of known symbols, a larger character set makes the task harder.
Additionally, because any kind of statistical language model needs adequate statistical representation of symbols in a corpus in order to handle them gracefully, the utility of singleton (or quasi-singleton) symbols is questionable.

In order to address the character set issue non-invasively, we propose an extensive use of charset simplification mappings throughout the Digital Humanities (DH) pipeline, so that the different use cases can rely on the character set that best suits them.

A closely related normalization problem is the treatment of abbreviations.
Medieval scribes abbreviated heavily, and whether to transcribe abbreviated or expanded forms is a long-standing dilemma with direct consequences for HTR training and downstream search.
Unlike charset simplification, expanding abbreviations is not a one-to-one operation---a single abbreviation mark may expand to several letters---so it requires modeling insertions and deletions, which we address with Banded RNNs.

The principal contributions of this paper are the following: we present \textit{pylelemmatize}, a library that efficiently implements charset simplification mappings (CSMs); we introduce letter lemmatization, a character similarity heuristic that automatically derives a CSM between any pair of character sets; we propose one-to-one RNNs that reverse a CSM through self-supervision, and use them both to obtain an empirical measurement of the information a CSM suppresses and as an HTR post-correction step; we introduce Banded RNNs, which extend the same architecture to mappings involving insertions and deletions, and apply them to medieval abbreviation expansion; finally, we make available a corpus for HTR post-correction with realistic, machine-generated errors, and a parallel corpus of Italian medieval abbreviations consisting of text-line pairs from 33 charters.

\section{Letter Lemmatization}
When considering alternatives in transcription policies
for medieval documents, we must consider how the resulting data will end up being used.
Document digitization already has a long tradition that goes back to the days when document images could not be processed digitally.
This has forged a mindset over the decades, where digital transcriptions are meant to be the sole computerized version of the document, and thus register all its content, with as many details as possible.
In this regard, digital transcriptions are to some extent analogs of copies made centuries ago in {\em scriptoria}, whose function was to facilitate the preservation and dissemination of documents.
The popularization of visual computing nowadays makes the close study of a manuscript without its image inconceivable.
In contemporary use cases, the role of a document transcription is shifting both towards facilitating the close reading in conjunction with the original image, and in an even larger measure toward automatic indexing of large corpora and distant reading analysis.
In short, transcriptions shift from being a digital representation a scholar can read to being something machines do read.
From a more utilitarian perspective, one could think of document transcription's principal purpose as being the groundtruth for training HTR models.
Given that a well-trained model can transcribe a set of documents that is several orders of magnitude larger than its train set, it essentially leverages the transcriber's effort by the same order of magnitude.
There are of course use cases, such as paleographic analysis, where a precise representation of ligatures, abbreviations, and similar features is critical: these specific applications do not lend themselves to simplifying the alphabet.

\subsection{Charset Simplification Mappings}
From an information perspective, a larger character set records more information.
We define a charset simplification mapping (CSM) as a one-to-one mapping from character to character, including a designated unknown character to which all out-of-vocabulary characters map.
With this in mind, we propose to employ CSMs as a thin software layer applied in real time rather than persisted to long-term storage, allowing one to ``tweak'' the character mapping at any stage of a pipeline without modifying the stored data.
In Python3 such behavior can be realized with a \textit{defaultdict}, but in essence a generic dictionary is all that is needed.
We thus present \textit{pylelemmatize}
\footnote{\textit{pylelemmatize} is available under the MIT license at~\url{https://github.com/anguelos/pylelemmatize}}
, a lightweight Python library that implements this functionality and many more such as unigram language model extraction, automatic character set alignment, and character error rate (CER) estimation.

\subsection{De-mapping}
\begin{figure}[t!]
  \centering
  \includegraphics[width=0.8\linewidth]{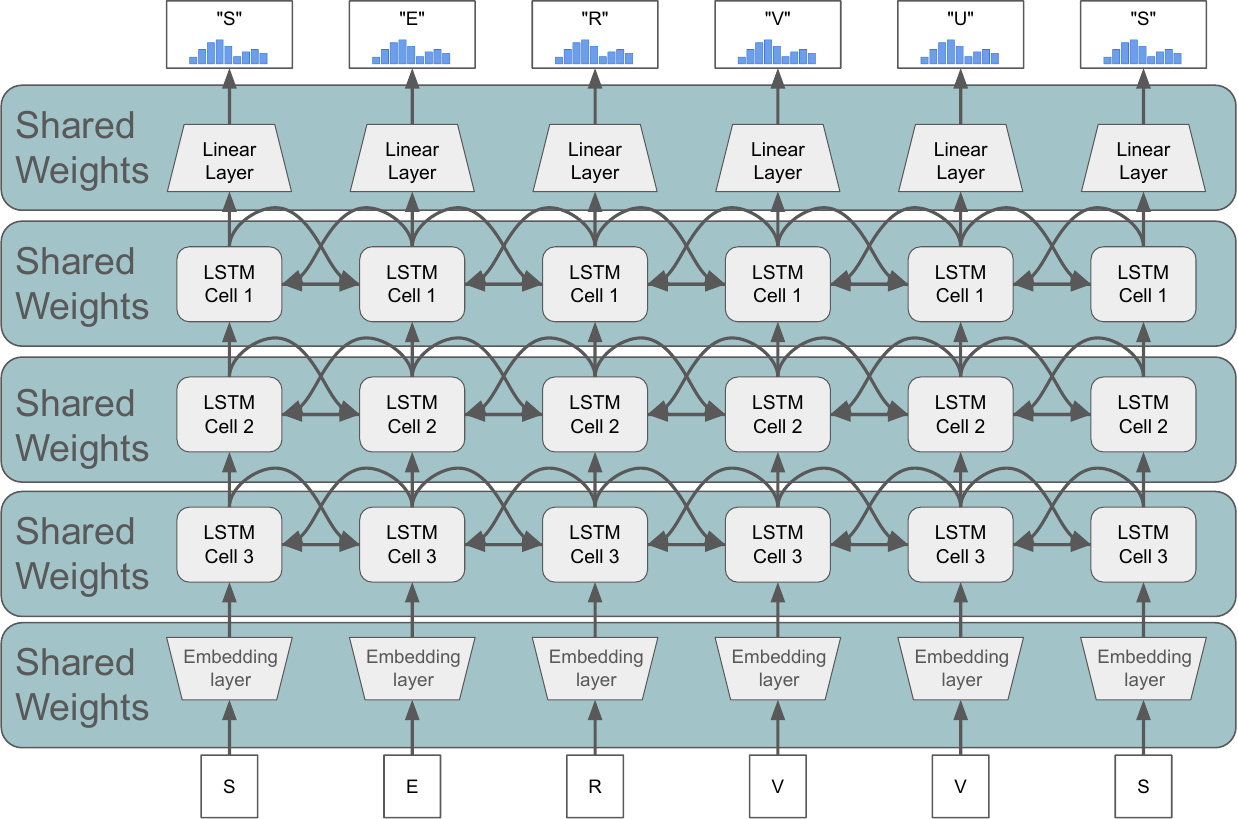}
  \caption{Architecture of a de-mapping (one-to-one) RNN, unrolled over the input ``SERVVS''.
  Each input character is embedded, passed through a stack of bidirectional LSTM layers with weights shared across positions, and projected by a per-position linear layer to a probability distribution over the destination characters.
  Input and output are aligned one-to-one, so the network is trained with per-position cross-entropy.}
  \label{fig:demapper}
\end{figure}

When more than one character maps to a single character, a CSM removes a bit of information every time it is applied.
Given a corpus, self-supervision can be used to learn to undo the mapping; this allows reducing the information loss such CSMs introduce.
Exploiting the one-to-one constraint of CSMs makes it much easier to train Recurrent Neural Networks (RNNs) such as Long-Short Term Memory (LSTM)~\cite{hochreiter1997long} networks, as inputs and outputs are perfectly aligned.
In Fig.~\ref{fig:demapper} the proposed architecture for learning how to reverse the mapping process can be seen.
The architecture has a single parameter for the size of the hidden layers, which is the same across the network, with the sole exception of the embedding layer, whose size also depends on the number of source characters, and the output layer, whose size depends on the number of destination characters.
As the context of both future and past events is important, bidirectional LSTM layers are employed.
The input for such networks is a sequence of integers, one for every character in the sequence being de-mapped, while the output is a probability distribution over each destination character for every input character.
We refer to these substitution-aligned networks as one-to-one RNNs.

\subsection{Letter Lemmatization}
\label{sec:letter_lemmatization}
Charset simplification mappings can be specified manually, but expressing a complete mapping for a large character set is laborious.
We introduce letter lemmatization: a heuristic that automatically derives a CSM between any pair of character sets by quantifying the similarity between characters.
The heuristic is specified in full below; a reference implementation is available with \textit{pylelemmatize}\footnote{Reference implementation: \url{https://github.com/anguelos/pylelemmatize/blob/main/src/pylelemmatize/char_distance.py}}.

The heuristic assigns a similarity of $1$ to identical characters.
For distinct characters it accumulates a score $s$ (initialised to $0$) from the weighted features below, where $[P]$ denotes $1$ if predicate $P$ holds and $0$ otherwise, $r(x,y)\in[0,1]$ is the \texttt{difflib} sequence-matching ratio $2M/(|x|+|y|)$ ($M$ = number of matched characters), and $u(\cdot)=\mathrm{unidecode}(\cdot)$ is the Unicode-to-ASCII transliteration~\cite{unidecode}:
\begin{itemize}
\item{\textbf{Unicode category:} $+0.1$ if the characters share the same major Unicode category (letter, number, punctuation, mark, symbol, \dots) and a further $+0.2$ if their full two-letter categories are identical.}
\item{\textbf{Unicode name overlap:} when both characters are named, with $T_a,T_b$ the sets of whitespace-separated tokens of their names $n_a,n_b$, add $0.3\,\frac{|T_a\cap T_b|}{\max(|T_a|,|T_b|)}+0.2\,r(n_a,n_b)+0.1\,r(n_a',n_b')$, where $n'$ is the first half of the name.}
\item{\textbf{Case, whitespace, punctuation:} $+0.2$ each if the characters are both alphabetic and equal up to case, both whitespace, or both ASCII punctuation.}
\item{\textbf{ASCII transliteration:} if $\max(|u(a)|,|u(b)|)>1$, add the position-weighted agreement $\frac{1}{|u(a)|\,|u(b)|}\sum_{i,j}[u(a)_i=u(b)_j]\,w_i\,w_j$, with weights $w$ decreasing linearly from $1.5$ (leftmost) to $0.5$ (rightmost) so that leading characters dominate; otherwise add $0.7\,[u(a)=u(b)]$.}
\item{\textbf{Case-insensitive transliteration:} the previous step is repeated on the lower-cased transliterations (single-character weight $0.5$ instead of $0.7$).}
\item{\textbf{Phonetic class:} using the lower-cased transliterations, add $0.25$ if they are equal, else $0.2$ if both are vowels, else $0.1$ if both are consonants.}
\item{\textbf{Ordinal proximity:} a minor term $0.05\,\sigma\!\left(|\mathrm{ord}(a)-\mathrm{ord}(b)|\right)$ with the logistic $\sigma(x)=1/(1+e^{-x})$.}
\end{itemize}
The accumulated score is squashed to $[0,1]$ by
\begin{equation}
f(s) = 2 \cdot \frac{1}{1 + e^{-s}} - 1,
\label{eq:scaled_sigmoid}
\end{equation}
and the final similarity is symmetrised as $\mathrm{sim}(a,b)=\tfrac12\big(f(s_{a,b})+f(s_{b,a})\big)$.

\begin{figure*}[t!]
  \centering
  \includegraphics[width=0.8\linewidth]{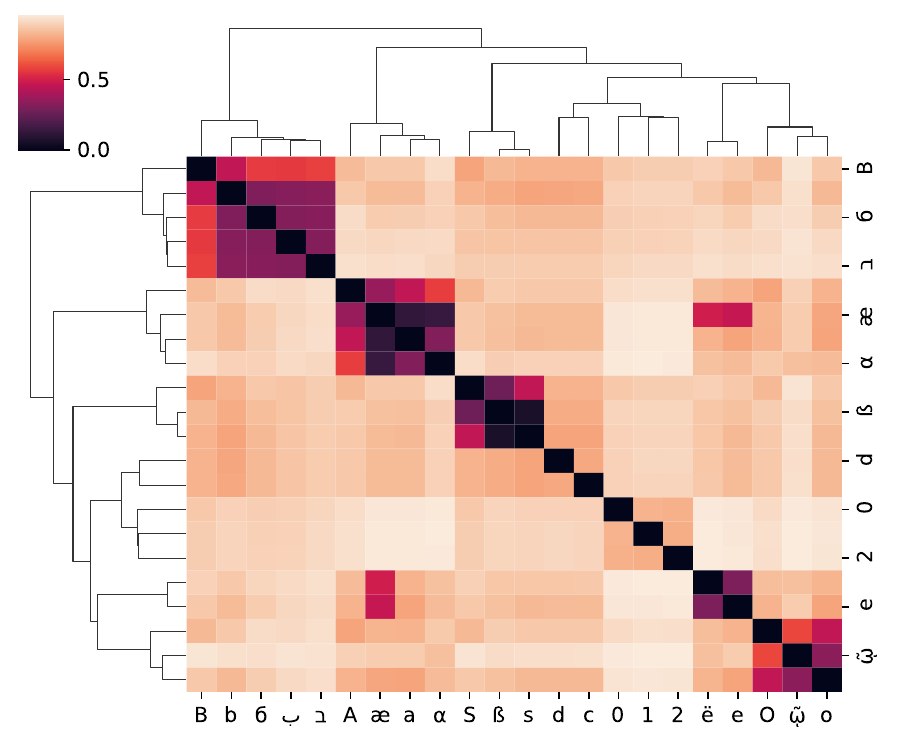}
  \caption{Indicative distance matrix of characters for letter lemmatization.}
  \label{fig:transliteration}

\end{figure*}

To build a CSM, each source character is mapped to the destination character with which it has the highest similarity; if that best similarity falls below $0.35$, the source character is mapped to the unknown character instead.
The mappings derived by letter lemmatization satisfy the one-to-one constraint and thus cannot elegantly handle multi-character mappings such as ligatures.
Fig.~\ref{fig:transliteration} shows an indicative distance matrix $(1 - similarity)$ between very diverse characters (from Latin, Cyrillic, Hebrew, Arabic, and Greek).

Mappings automatically derived by letter lemmatization can easily be overridden in \textit{pylelemmatize} but they offer a convenient scaffold even when they are not the best for a specific use-case.

\subsection{Banded RNNs}
\label{sec:banded_rnn}
One-to-one RNNs exploit the substitution-only constraint: input and output sequences are perfectly aligned, allowing cross-entropy training at every position.
When a normalization involves insertions or deletions---as in abbreviation expansion, where a single abbreviated character may expand to several letters---this constraint cannot be satisfied.
We extend the same bidirectional LSTM architecture to this setting by modifying only the dataset construction and the training loss; the network itself is unchanged.

Given a parallel corpus, dynamic programming is used to align each source character to a contiguous output sequence of length zero to $N$.
Text lines in which any source character maps to more than $N$ output characters are discarded.
Each source character is then replicated $N$ times before being fed to the RNN, giving the Connectionist Temporal Classification (CTC) decoder~\cite{graves2006connectionist} room to emit up to $N$ output characters per input position.
The one-to-one RNN is the $N=1$ special case of this framework, where cross-entropy replaces CTC.
\begin{table*}[!t]
    \centering
\caption{pylelemmatize computation speed vs. baselines}
\scriptsize
\begin{tabular}{ccc||c|c|c|c}
\hline
\multicolumn{3}{c||}{Experiment Parameters} & \multicolumn{4}{c}{Bench-marked Methods (msec.)} \\
\hline
\makecell{Sample\\Size (KB)} &
\makecell{Unknown\\Characters (\%)} &
\makecell{Mapping\\Size} &
\makecell{Python\\dict} &
\makecell{Python\\defaultdict} &
\makecell{Generic\\Lemmatizer} &
\makecell{Fast\\Lemmatizer} \\
\hline
1 & 0.0 & 4  &  $ 47.4 \pm 2.0 $ &  $ 37.4 \pm 1.3 $ &  $ 101.4 \pm 4.0 $ &\boldmath{ $ 13.1 \pm 0.4 $} \\
1 & 0.0 & 3549  &  $ 50.1 \pm 1.6 $ &  $ 42.1 \pm 2.7 $ &  $ 102.8 \pm 3.0 $ &\boldmath{ $ 13.0 \pm 0.5 $} \\
10000 & 0.0 & 4  &  $ 49.9 \pm 0.9 $ &  $ 39.6 \pm 0.6 $ &  $ 104.8 \pm 3.2 $ &\boldmath{ $ 6.7 \pm 0.1 $} \\
10000 & 0.0 & 3549  &  $ 50.7 \pm 0.9 $ &  $ 39.0 \pm 0.3 $ &  $ 100.5 \pm 0.9 $ &\boldmath{ $ 7.2 \pm 0.5 $} \\
10000 & 0.01 & 4  &  $ 51.2 \pm 0.1 $ &  $ 39.3 \pm 0.4 $ &  $ 106.8 \pm 0.7 $ &\boldmath{ $ 6.5 \pm 0.2 $} \\
10000 & 0.01 & 3549  &  $ 53.9 \pm 2.6 $ &  $ 38.8 \pm 0.1 $ &  $ 108.4 \pm 2.2 $ &\boldmath{ $ 6.5 \pm 0.0 $} \\
10000 & 1.0 & 4  &  $ 53.9 \pm 0.3 $ &  $ 38.8 \pm 0.2 $ &  $ 107.0 \pm 2.6 $ &\boldmath{ $ 7.7 \pm 0.3 $} \\
10000 & 1.0 & 3549  &  $ 59.0 \pm 0.3 $ &  $ 39.4 \pm 1.0 $ &  $ 111.1 \pm 1.8 $ &\boldmath{ $ 7.5 \pm 0.1 $} \\
\hline
\end{tabular}
\label{tab:benchmark}
\end{table*}

\section{Experiments}
For experiments, we chose the Königsfelden~\cite{koeningsfelden} corpus, an ideal fit to our intended use cases.
Specifically it consists of \numprint{3102} text lines, \numprint{71235} words, and \numprint{448073} characters.
In the default case, the corpus is partitioned at the text line level, with 80\% of the lines assigned to the train set and 20\% to the validation set.
When we refer to a lemmatizer with a character set's name, e.g. the MUFI lemmatizer, we refer specifically to the subset of characters that is inside the BMP and outside of the PUA regions, while mapping any out-of-set character to  the Unicode \textsc{replacement character}.
There are a few characters observed in the Königsfelden corpus that are not even part of MUFI---for instance, the character \textsc{eight Spoked asterisk} is frequently used there to mark where an abbreviation was expanded.

\subsection{Benchmarking pylelemmatize}
In order to optimize performance, \textit{pylelemmatize} was built on top of the Numpy library instead of the generic Python tools.
We ran benchmarks comparing \textit{pylelemmatize} to generic implementations of one-to-one mapping done with Python.
As can be seen in Table~\ref{tab:benchmark}, the \textit{pylelemmatize} library is quite faster than generic mapping methods, as long as all characters reside in the Unicode Basic Multilingual Plane (BMP) range (marked as Fast Lemmatizer).
The benchmark tests parameters such as the size of the mapped sample, the size of the mapping, the frequency of unknown characters in the sample.
Compared to standard Python tools, \textit{pylelemmatize} brings a speed-up between $2.8 \times $ and $6 \times$ depending on the parameters.
\textit{Chokomufin}~\cite{Clerice_Choco-Mufin_a_tool_2021} has capabilities \textit{pylelemmatize} has not, as it addresses a more general case of transformations, yet for the specific task of extracting observed alphabets from a corpus, a comparison to the \textit{pylelemmatize} demo tool \textit{ll\_extract\_corpus\_alphabet} makes sense.
In this case, \textit{ll\_extract\_corpus\_alphabet} runs in 0.247 seconds on a corpus of 2.99 MB, instead of 0.541 sec for \textit{Chokomufin}'s \textit{generate} utility.

\begin{table*}[!t]
\centering
\caption{De-mapping recovery rates}
\scriptsize
\label{tab:demapping}
\begin{tabular}{cccc||ccc}
\hline
\makecell{Source\\Alphabet} & \makecell{Destination\\Alphabet} &  \makecell{Override\\Mapping} & Corpus &\makecell{Mapping\\CER (\%)} & \makecell{De-mapping\\CER (\%)} &  Recovery (\%) \\
\hline
MUFI(BMP) & ASCII(lower) & - & Kngsfldn & 4.04 & 0.96 & 76.12 \\
MUFI(BMP) & ASCII(mini) & - & Kngsfldn & 4.85 & 1. & 79.38 \\
MUFI(BMP) & MUFI(BMP) & U$\to$ V & Kngsfldn & 5.08 & 0.01 & 99.99 \\
\hline
ASCII & \makecell{ASCII \\consonants 25} & \makecell{AEIOU $\to$ A\\0123456789 $\to$ 1\\punctuation $\to$ .} &\makecell{tiny\\shakespeare\\ \cite{tinyshakepare}} & 33.39 & 4.65 & 86.07\\
\hline
\end{tabular}
\end{table*}

\subsection{De-mapping}
We performed some indicative experiments involving self-supervised learning on RNNs that learn to reverse CSMs, in order to quantify the information loss that results from applying a specific CSM.
The results can be seen in Table~\ref{tab:demapping}\footnote{The experiments can be reproduced and their details understood in \url{https://github.com/anguelos/pylelemmatize/tree/main/experiments/delemmatization}}.
Our experiments consist of applying an automatic mapping from \textit{Source Alphabet} to \textit{Destination Alphabet}, while adding in some cases more arbitrary character mappings (\textit{Override Mapping}).
We randomly split our corpus text lines into 80\% for a train set and 20\% for a validation set.
For every CSM we measure the character error rate (CER) between the test set and the mapped test set (\textit{mapping CER}) and we then measure the CER between the test set and the de-mapper applied on the mapped test set.
When mapping MUFI (777 characters) to the ASCII Lowercase (74 characters), we show that in the worst case scenario, reducing the output layer of an HTR neural network from 777 neurons to 74 would increase the CER by less than 0.96\%.
Our mapping of MUFI to ASCII Mini (39 characters) suggests that reducing the output layer to 39 neurons would be marginally worse at 1.0\%.
At the same time the U $\to$ V experiment indicates that such a CSM is totally recoverable, indicating that we can rectify corpora that have been inconsistently transcribed with U or V and could become consistent by applying a U $\to$ V CSM, followed by de-mapping trained on a consistent sub-corpus.
In this regard, the small Shakespeare experiment, that provides quantitative insights into the information loss of applying consonantal writing to Elizabethan English, has a rather limited scientific merit, but demonstrates nonetheless the flexibility of \textit{pylelemmatize}.

\subsection{HTR post-correction}
The de-mapping RNNs can be employed in any one-to-one mapping of sequences.
HTR errors consist of insertions, deletions, and typically mostly substitutions.
We performed a simple experiment to see whether we could learn how to correct any substitution errors by using our de-mapping RNNs.

\begin{table}[]
    \caption{HTR Post-correction corpora}
    \small
    \centering
    \begin{tabular}{l|ccc}
    \hline
        Name & Lines & Tokens & Chars. \\
    \hline
        Monasterium &  \numprint{11168} & \numprint{196526} & \numprint{1246449} \\
         Königsfelden  &  \numprint{26117} & \numprint{440520} & \numprint{2628389} \\
         Nurn. basic   &  \numprint{50895} & \numprint{402797} & \numprint{2412584} \\
         Nurn. norm.   &  \numprint{26129} & \numprint{212342} & \numprint{1266254} \\
         Total & \numprint{114309} & \numprint{1252185} & \numprint{7553676} \\
    \hline
    \end{tabular}
    \label{tab:corpora}
\end{table}

\subsubsection{Corpora}
We curated four textline-aligned parallel corpora from HTR-training datasets by having an HTR-engine based on kraken~\cite{kiessling2019kraken} run on them.
In Table~\ref{tab:corpora} we can see the cardinalities of the corpora.
The corpora are stored as Unicode \textit{.tsv} files where the first column is the noisy HTR prediction and the second is the groundtruth\footnote{The corpus will be made available upon publication.}.
The employed corpora are derivatives of \textit{Königsfelden}~\cite{koeningsfelden}, and the \textit{Nuremberg Letterbooks}~\cite{mayr2025nuremberg}.
The \textit{Monasterium} corpus has been curated from scratch; it has been collected from the same source as the corpus presented in~\cite{atzenhofer2024text}, so there might be partial overlap of the documents.
The corpus was built as follows: (1) From those charters that are equipped with a full, page-wide transcription, we extract the tenor---because the work has been done over time by scholars with different skills and goals, the transcription style varies significantly between documents; most documents are in German ($\approx$60\%) or Latin ($\approx$39\%), with a small number of Czech or Hungarian manuscripts. %

\subsubsection{Experimental Procedure}
For every sub-corpus, we shuffled the text-lines randomly, sequestered \numprint{2000} rows as a test-set, and kept the rest as an experimental set.
We filtered out lines with a CER greater than 20\% or shorter than 50 characters in both test and experimental partitions assuming they are miss-segmented or problematic text-lines.
For the experimental data partition, for every line, we selected substitution-only errors and aligned them to the groundtruth thus making pairs of text-lines with exactly the same length such that the substitution-only line and the groundtruth string have no insertions or deletions when compared.
We then split the experimental partition randomly to 80\% for a train-set and 20\% for a validation-set and train the exact same RNN previously described for de-mapping.

\begin{table}[]
    \centering
    \caption{Post-HTR correction performance}
    \begin{tabular}{l||cc|cc}
    \hline
          & \multicolumn{2}{c|}{val. CER \%} & \multicolumn{2}{c}{Test CER \%}  \\
          \hline
         Corpus & HTR & Corr. & HTR & Corr. \\

         \hline
         Monasterium  & 6.66 & 5.74 & 7.25  & 7.19 \\
         Königsfelden   & 5.11  & 3.59    & 5.78  & 4.58 \\
         Nurn. basic     & 17.26 & 6.20   & 18.57 & 8.18 \\
         Nurn. norm.    & 6.81 & 3.88   &  8.05 & 5.87 \\
         All train-test  & 10.16 & 5.52 & 8.83 & 6.60  \\
         \hline
    \end{tabular}

    \label{tab:postcorrection}
\end{table}

\subsubsection{Results}
In Table~\ref{tab:postcorrection} the results of the post-correction experiment can be seen.
The validation CER is measured on the substitution-only inputs that were created by aligning at the character-level HTR outputs to the groundtruth preserving only substitution errors.
Test CER on the other hand is measured by employing the trained RNN as intended in real-world use-cases where the HTR outputs are passed through the RNN as-is and we then measure empirically its CER compared to the groundtruth; this is a better predictor of the performance improvements we can expect on HTR output.
As expected since the majority of errors occurring in the test-set are substitutions, validation CER generalizes quite well in the test-set.
From the four corpora employed, it stands out that the Monasterium sub-corpus demonstrates no meaningful improvement for correcting HTR outputs, we speculate the reasons for this are many, including the inconsistent transcription practices, the multilingual nature of the corpus, the relatively small corpus size.
Finally the \textit{all} experiment in which we concatenated all experimental partitions to train a single model and then used it to correct all test lines, demonstrates that a single model can be used on heterogeneous data.

\subsection{Abbreviation Expansion with Banded RNNs}
\label{sec:abbrev_expansion}
Medieval scribes routinely abbreviated words, and the choice of whether to transcribe abbreviated or expanded forms has significant consequences for downstream pipelines.
We demonstrate Banded RNNs on this task using two parallel corpora of text lines aligned at the character level between abbreviated and expanded forms.

\subsubsection{Corpora}
The Fontenay corpus (FTN)~\cite{camps2022dh,camps2021handlingheavilyabbreviatedmanuscripts} provides parallel abbreviated and manually expanded transcriptions of medieval manuscripts.
The Santa Maria della Grotta corpus (SMG) is a new corpus of \numprint{33} charters from the former monastery of Santa Maria della Grotta in Vitulano (BN, Italy), now held at the Societ\`a Napoletana di Storia Patria, dated 1254--1380 and issued across the Kingdom of Naples.
Both corpora are stored as text-line-aligned \textit{.tsv} files pairing each abbreviated line with its manually expanded counterpart.
Key statistics are shown in Table~\ref{tab:abbrev_corpora}.
The CER between abbreviated and expanded forms is \numprint{21.6}\% for FTN and \numprint{31.1}\% for SMG, reflecting how aggressively each corpus abbreviates.
Abbreviated texts use substantially larger character sets than their expanded counterparts, as special abbreviation characters are replaced by sequences of common letters during expansion.

\begin{table}[t]
    \centering
    \caption{Abbreviation expansion corpora}
    \label{tab:abbrev_corpora}
    \small
    \begin{tabular}{l|cc|cc}
    \hline
     & \multicolumn{2}{c|}{Charset Size} & \multicolumn{2}{c}{Length (chars)} \\
    Corpus & Abr. & Exp. & Abr. & Exp. \\
    \hline
    SMG & 104 & 58 & \numprint{250763} & \numprint{293445} \\
    FTN & 81  & 36 & \numprint{250689} & \numprint{269561} \\
    \hline
    \end{tabular}
\end{table}

\subsubsection{Experimental Procedure}
For each corpus, text lines were aligned at the character level using dynamic programming; lines in which any single source character maps to more than $N=5$ target characters were discarded from both the training and evaluation partitions.
We fixed $N=5$ rather than tuning it: individual abbreviations typically expand to one to three characters, and the expansion-length distribution is heavy-tailed (roughly Pareto), so a band of five is a safety margin that already covers the large majority of cases while keeping the padded inputs compact.
A larger band would admit only rare long-tail expansions at the cost of wider inputs and slower convergence; since both partitions apply the same $N=5$ filter, the reported CERs are conditional on expansions that fit within the band, and a systematic sweep of $N$ is left for future work.
We trained one Banded RNN per corpus in the expansion direction (abbreviated $\to$ expanded) and the contraction direction (expanded $\to$ abbreviated).
We also trained a single mixed model on the union of both corpora to assess cross-corpus generalisation.
All models use the same bidirectional LSTM architecture as the de-mapping RNNs.

\subsubsection{Results}
\label{sec:abbrev_results}
The results are shown in Table~\ref{tab:abbrev_results}.
As a trivial no-op baseline we take the identity mapping CER---the error a downstream consumer incurs by leaving abbreviations unexpanded---which is \numprint{21.6}\% for FTN and \numprint{31.1}\% for SMG.
Expansion (abbreviated $\to$ expanded) achieves \numprint{3.9}\% CER on FTN and \numprint{4.2}\% CER on SMG, reducing this no-op baseline by roughly $5$--$7\times$.
The reverse transformation (expanded $\to$ abbreviated) performs significantly worse: \numprint{16.4}\% CER for FTN and \numprint{10.3}\% for SMG.
This asymmetry indicates that abbreviation expansion is a largely irreversible transformation: a given expanded sequence can correspond to multiple valid abbreviated forms, making the contraction direction less determined.
The mixed model trained on the union of both corpora achieves \numprint{4.3}\% CER against a \numprint{26.5}\% no-op baseline, demonstrating that a single Banded RNN generalises well across corpora with differing transcription practices.

\begin{table}[t]
    \centering
    \caption{Abbreviation expansion and contraction CER (\%).
    The \emph{no-op baseline} is the trivial identity mapping CER---the error incurred by leaving the text unexpanded---and is a lower reference bar rather than a competing method.
    The SMG+FTN no-op baseline is the length-weighted combination of the two corpus no-op baselines.}
    \label{tab:abbrev_results}
    \small
    \begin{tabular}{l||cc|c}
    \hline
     & \multicolumn{2}{c|}{Expansion CER (\%)} & Contraction \\
    Corpus & No-op & Banded RNN & CER (\%) \\
    \hline
    SMG      & 31.1 & 4.2 & 10.3 \\
    FTN      & 21.6 & 3.9 & 16.4 \\
    SMG+FTN  & 26.5 & 4.3 & --   \\
    \hline
    \end{tabular}
\end{table}

\subsection{De-mapping ablation study}
\begin{figure}[h!]
    \centering
    \includegraphics[width=0.9\linewidth]{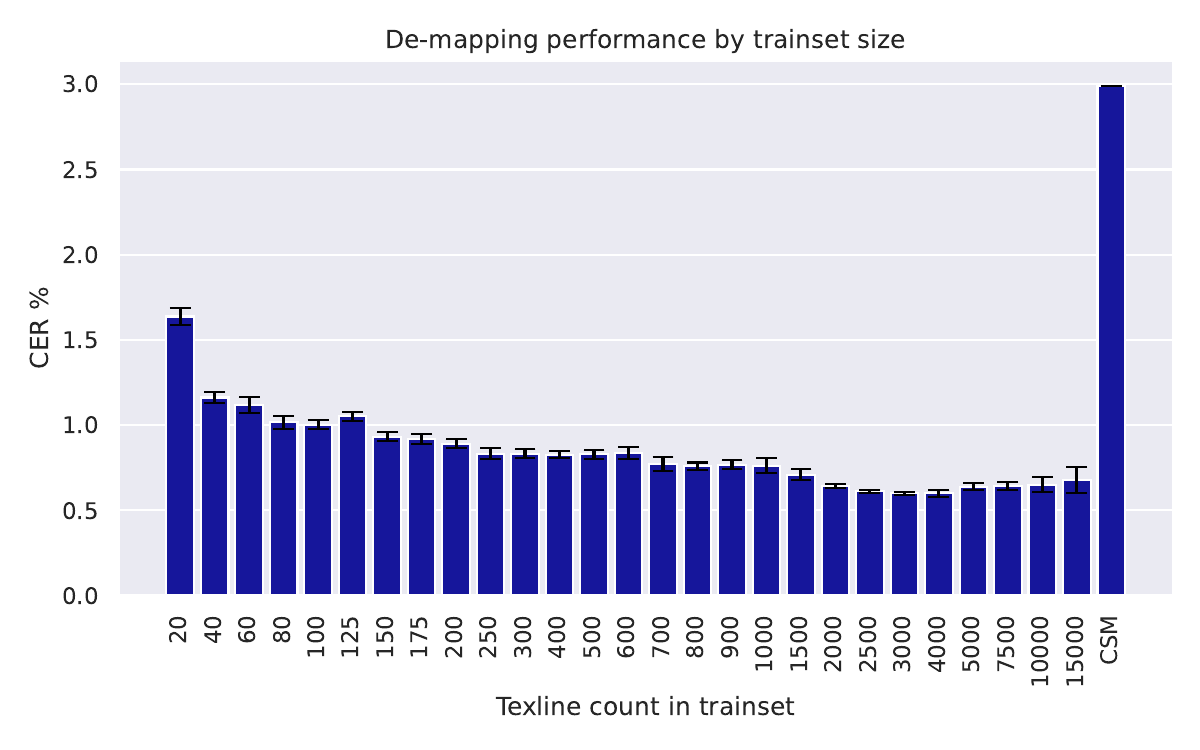}
    \caption{De-mapping CER as a function of training set size (MUFI to ASCII CSM, $110 \rightarrow 72$ characters, Königsfelden corpus).
    Error bars reflect training dynamics; the horizontal reference marks the original CSM CER.
    De-mapping halves the mapping error from as few as 20 training lines, with gains plateauing near 4,000 lines.}
    \label{fig:ablation}
\end{figure}

We performed an ablation study to provide insights on the data requirements of de-mapping.
We trained de-mapping RNNs to undo the CSM of all occurring characters of the MUFI character set (110 occurring characters) to an ASCII subset (72 occurring characters).
We used a typical LSTM with three layers of 256 recurrent neurons.
We fixed the validation set to \numprint{4316} text lines and then ran multiple training sessions, progressively increasing the train set size from 20 text lines to \numprint{15000} text lines.
All training sessions applied approximately the same number of iterations, about \numprint{30000}.
In Fig.~\ref{fig:ablation} the character error rate (CER) of each de-mapper can be seen, with the original CSM CER also provided as a reference.
In our initial runs, we observed that de-mapping RNNs do not exhibit overfitting, so that an early stopping validation set is not really needed; we just train them for longer than we estimate necessary and measure the validation performance at the last epoch.
Furthermore, under the assumption that during the second half of the training the models have already converged, we use the standard deviation of validation errors during the second half of the training as error bars in the plot.
The error bars do not account for the random initialization of the networks, but they demonstrate the randomness that can be attributed to training dynamics.
It can be seen that even from a train-set of 20 text lines, de-mapping can halve the mapping error; after 4,000 text-lines there seems to be no more improvement for our corpus.

\section{Conclusions}

\subsection{Discussion}
This work was initially motivated by the need to understand empirically the tradeoff between reducing the alphabet size and preserving as much information as possible when devising an HTR policy or transcription guidelines.
Many experiments in this paper were also done to demonstrate the benefits the one-to-one approach brings and how \textit{pylelemmatize} facilitates experimenting.
Although the corpus used for this exploration is quite specific---it is a large collection of Central European medieval charters---we now appreciate that techniques such as letter lemmatization and de-mapping RNNs can be just as useful for any task involving text that is noisy, or borne out of an aggregation of heterogeneous sources, with varying encoding practices.
The one-to-one RNNs are easy to train even with a surprisingly small amount of data: this suggests that we should favor a substitution-only model for as many tasks as possible.
In many cases text normalization schemes could be realized only with one-to-one operations.
A major benefit of the one-to-one constraint is that it is also friendly to standoff annotations, as it does not affect them at all, and can thus be employed inside complicated pipelines with minor modifications.
At the same time, in the case of inline annotations such as in XML, if the mapper maps all the structure characters to themselves, a CSM should preserve the structure of any string it maps.
Because a CSM introduces no insertions or deletions, computing the CER between a string and its mapped counterpart requires no dynamic-programming alignment; it reduces to a position-wise character comparison, which streamlines many experimental measurements.

\subsection{Limitations}
The one-to-one mapping constraint limits the quality of the resulting mappings, particularly for ligatures and abbreviations, and de-mapping RNN training needs a corpus that adequately represents all mapping operations and their context.
Where the character-similarity heuristic does not realize an intended mapping, the user must define and maintain a custom one.
Banded RNNs lift the one-to-one constraint but at a cost: the band size $N$ bounds how far a single character may expand, so lines whose alignment exceeds the band must be re-mapped or discarded, and an ill-chosen $N$ either wastes data or inflates the inputs.
Banded RNNs also converge more slowly and need considerably more data than the remarkably data-efficient one-to-one de-mappers, trading that efficiency for the ability to model insertions and deletions.

\subsection{Future Work}
The Banded RNNs introduced here already relax the one-to-one constraint to handle insertions and deletions; extending this to full many-to-many mappings would allow modeling ligatures and more complex abbreviations, likely requiring larger corpora.
We are working on extending letter lemmatization to what we call letter tokenization.
Having shown that de-mapping RNNs are effective for HTR post-correction, we intend to examine experimentally whether reducing the HTR character set and applying de-mapping RNNs as a post-correction step could improve end-to-end HTR performance.
Recent work applies fine-tuned transformer and large-language-model post-correction to HTR and to abbreviation expansion; such models are a natural strong baseline, but they rely on substantial pretraining and data, whereas our character-level RNNs target a complementary low-resource, lightweight niche---recovering half the CER from as few as 20 lines and running as a thin layer over existing pipelines.
A systematic comparison against transformer-based post-correction is left for future work.
While our experiments gave good results on our limited corpora, it would be interesting to perform them on more diverse corpora.


\begin{thebibliography}{10}
\providecommand{\url}[1]{\texttt{#1}}
\providecommand{\urlprefix}{URL }
\providecommand{\doi}[1]{https://doi.org/#1}

\bibitem{atzenhofer2024text}
Atzenhofer-Baumgartner, F., Kov{\'a}cs, T.: Is text normalization relevant for
  classifying medieval charters? In: International Conference on Theory and
  Practice of Digital Libraries. pp. 125--132. Springer (2024)

\bibitem{bollmann-2019-large}
Bollmann, M.: A large-scale comparison of historical text normalization
  systems. In: Burstein, J., Doran, C., Solorio, T. (eds.) Proceedings of the
  2019 Conference of the North {A}merican Chapter of the Association for
  Computational Linguistics: Human Language Technologies, Volume 1 (Long and
  Short Papers). pp. 3885--3898. Association for Computational Linguistics,
  Minneapolis, Minnesota (Jun 2019). \doi{10.18653/v1/N19-1389},
  \url{https://aclanthology.org/N19-1389}

\bibitem{breuel2008ocropus}
Breuel, T.M.: The ocropus open source ocr system. In: Document recognition and
  retrieval XV. vol.~6815, pp. 120--134. SPIE (2008)

\bibitem{breuel2013high}
Breuel, T.M., Ul-Hasan, A., Al-Azawi, M.A., Shafait, F.: High-performance ocr
  for printed english and fraktur using lstm networks. In: 2013 12th
  international conference on document analysis and recognition. pp. 683--687.
  IEEE (2013)

\bibitem{unidecode}
Burke, S.M., Šolc, T.: unidecode. \url{https://github.com/avian2/unidecode}
  (2025), gitHub repository, commit \texttt{8d83b7c}

\bibitem{camps2022dh}
Camps, J.B., Vidal-Gor\`ene, C., Stutzmann, D., Vernet, M., Pinche, A.: Data
  diversity in handwritten text recognition: challenge or opportunity? In:
  Digital Humanities 2022: Conference Abstracts. pp. 160--165. DH2022 Local
  Organizing Committee, Tokyo (2022),
  \url{https://dh2022.dhii.asia/dh2022bookofabsts.pdf}

\bibitem{camps2021handlingheavilyabbreviatedmanuscripts}
Camps, J.B., Vidal-Gorène, C., Vernet, M.: Handling heavily abbreviated
  manuscripts: Htr engines vs text normalisation approaches (2021),
  \url{https://arxiv.org/abs/2107.03450}

\bibitem{Clerice_Choco-Mufin_a_tool_2021}
Clérice, T., Pinche, A.: {Choco-Mufin, a tool for controlling characters used
  in OCR and HTR projects} (Sep 2021). \doi{10.5281/zenodo.5356154},
  \url{https://github.com/PonteIneptique/choco-mufin}

\bibitem{graves2006connectionist}
Graves, A., Fern{\'a}ndez, S., Gomez, F., Schmidhuber, J.: Connectionist
  temporal classification: labelling unsegmented sequence data with recurrent
  neural networks. In: Proceedings of the 23rd international conference on
  Machine learning. pp. 369--376 (2006)

\bibitem{koeningsfelden}
Halter-Pernet, C., Teuscher, S., Hodel, T., Barwitzki, L., Egloff, S.,
  Henggeler, F., Nadig, M., Steinmann, A., Stettler, S., Prada~Ziegler, I.:
  Charters and records of königsfelden abbey and bailiwick (1308-1662) (Aug
  2021). \doi{10.5281/zenodo.5179361},
  \url{https://doi.org/10.5281/zenodo.5179361}

\bibitem{haugen2009mufi}
Haugen, O.E.: Mufi character recommendation v. 3.0 (2009)

\bibitem{hochreiter1997long}
Hochreiter, S., Schmidhuber, J.: Long short-term memory. Neural computation
  \textbf{9}(8),  1735--1780 (1997)

\bibitem{tinyshakepare}
Karpathy, A.: char-rnn. \url{https://github.com/karpathy/char-rnn} (2015)

\bibitem{kiessling2019kraken}
Kiessling, B.: Kraken-a universal text recognizer for the humanities. In:
  Digital Humanities 2019 (2019)

\bibitem{kragl2015}
Kragl, F.: Normalmittelhochdeutsch. theorieentwurf einer gelebten praxis.
  Zeitschrift für deutsches Altertum und deutsche Literatur  \textbf{144}(1),
  1--27 (2015), \url{http://www.jstor.org/stable/43672053}

\bibitem{mayr2025nuremberg}
Mayr, M., Krenz, J., Neumeier, K., Bub, A., Bürcky, S., Brolich, N., Herbers,
  K., Habermann, M., Fleischmann, P., Maier, A., Christlein, V.: Nuremberg
  letterbooks: A multi-transcriptional dataset of early 15th century
  manuscripts for document analysis. Scientific Data  \textbf{12}, ~811 (May
  2025). \doi{10.1038/s41597-025-05144-z},
  \url{https://doi.org/10.1038/s41597-025-05144-z}, published online 17 May
  2025

\bibitem{distantReading_moretti2000conjectures}
Moretti, F.: Conjectures on world literature. New left review  \textbf{2}(1),
  54--68 (2000)

\bibitem{ocrd}
Neudecker, C., Baierer, K., Federbusch, M., Boenig, M., W\"{u}rzner, K.M.,
  Hartmann, V., Herrmann, E.: Ocr-d: An end-to-end open source ocr framework
  for historical printed documents. In: Proceedings of the 3rd International
  Conference on Digital Access to Textual Cultural Heritage. p. 53–58.
  DATeCH2019, Association for Computing Machinery, New York, NY, USA (2019).
  \doi{10.1145/3322905.3322917}, \url{https://doi.org/10.1145/3322905.3322917}

\bibitem{icdar2017_HTRcompetition}
Sanchez, J.A., Romero, V., Toselli, A.H., Villegas, M., Vidal, E.: Icdar2017
  competition on handwritten text recognition on the read dataset. In: 2017
  14th IAPR international conference on document analysis and recognition
  (ICDAR). vol.~1, pp. 1383--1388. IEEE (2017)

\bibitem{icdar2015_HTRcompetition}
S{\'a}nchez, J.A., Toselli, A.H., Romero, V., Vidal, E.: Icdar 2015 competition
  htrts: Handwritten text recognition on the transcriptorium dataset. In: 2015
  13th International Conference on Document Analysis and Recognition (ICDAR).
  pp. 1166--1170. IEEE (2015)

\bibitem{smith2007overview}
Smith, R.: An overview of the tesseract ocr engine. In: Ninth international
  conference on document analysis and recognition (ICDAR 2007). vol.~2, pp.
  629--633. IEEE (2007)

\bibitem{unicode2021}
{The Unicode Consortium}: The Unicode Standard, Version 14.0.0. The Unicode
  Consortium, Mountain View, CA (Sep 2021)

\end{thebibliography}
\end{document}